\documentclass[preprint,12pt]{elsarticle}

\usepackage{amssymb}
\usepackage{amsmath}

\usepackage{lineno}
\usepackage{booktabs}

\usepackage{graphicx}

\usepackage{amsmath,amssymb} 
\usepackage{rotating}

\usepackage{multirow}
\usepackage[usenames,dvipsnames]{pstricks}
\usepackage{url}
\usepackage{enumitem}

\newcommand{\p}{\mathbf{p}}

\renewcommand{\r}{\mathbf{r}}

\newcommand{\s}{\mathbf{s}}
\renewcommand{\t}{\mathbf{t}}

\newcommand{\x}{\mathbf{x}}

\newcommand{\mK}{\mathtt{K}}

\newcommand{\mP}{\mathtt{P}}

\newcommand{\mR}{\mathtt{R}}

\newcommand{\Comment}[1]{}

\journal{Computer Vision and Image Understanding}

\begin{document}

\begin{frontmatter}



\title{Unsupervised Monocular Road Segmentation for Autonomous Driving via Scene Geometry}


\author[label1]{Sara Hatami Rostami}
\ead{s.hatamirostami@email.kntu.ac.ir}

\author[label1]{Behrooz Nasihatkon}
\ead{nasihatkon@kntu.ac.ir}

\affiliation[label1]{organization={K.N. Toosi University of Technology},
	addressline={Department of Electrical and Computer Engineering}, 
	city={Tehran},
	postcode={16317-14191}, 
	state={Tehran},
	country={Iran}}


%

\begin{abstract}
This paper presents a fully unsupervised approach for binary
road segmentation (road vs. non-road), eliminating the
reliance on costly manually labeled datasets. The method
leverages scene geometry and temporal cues to distinguish
road from non-road regions. Weak labels are first generated
from geometric priors, marking pixels above the horizon as
\emph{non-road} and a predefined quadrilateral in front of the vehicle
as \emph{road}. In a refinement stage, temporal consistency is
enforced by tracking local feature points across frames and
penalizing inconsistent label assignments using mutual information
maximization. This enhances both precision and
temporal stability. On the Cityscapes dataset, the model
achieves an Intersection-over-Union (IoU) of 0.86, outperforming the competing unsupervised methods. These findings
demonstrate the potential of combining geometric constraints
and temporal consistency for scalable unsupervised
road segmentation in autonomous driving.
\end{abstract}




\end{frontmatter}



\section{Introduction}
\label{Introduction}
\begin{figure}[t]
	\centering
	\includegraphics[width=1\linewidth]{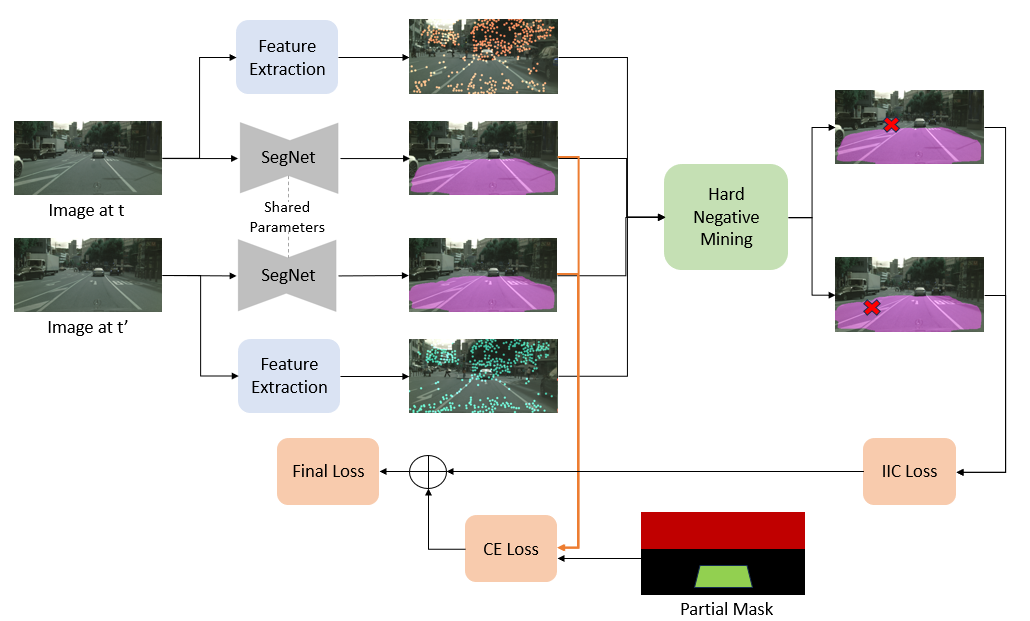}
	\caption{Overall framework of the proposed approach: initial training with partial mask and subsequent refinement with feature extraction and temporal consistency constraints. Feature extraction consists of local feature detection and tracking across frames. The tracked features are then filtered through hard negative mining before being fed into the IIC loss \citep{Ji2019IIC}. Finally, the IIC loss is combined with the cross-entropy loss to form the final loss for training the segmentation network.}
	\label{figone}
\end{figure}

Road detection and segmentation are among the most vital tasks in autonomous 
driving. Semantic segmentation serves as a major tool for this purpose, achieving high performance when trained in a supervised manner by leveraging densely annotated datasets~\citep{long2015FCN, Ronneberger2015U-net}. However, supervised semantic segmentation approaches rely heavily on pixel-wise annotations, which are costly and time-consuming to obtain~\cite{Chen2017Deeplab}. To address the challenges of expensive and labor-intensive labeling, semi-supervised, weakly supervised, and unsupervised approaches have been explored. This study evaluates the effectiveness of a fully unsupervised approach for semantic segmentation of road scenes. While unsupervised methods are typically complemented by other approaches in safe autonomous driving pipelines, our aim here is to assess how well a fully unsupervised approach can perform on its own. 

We propose a simple yet effective framework for processing road images captured by a calibrated camera mounted on a moving vehicle. The method classifies image pixels into road and non-road categories. In the first stage, a preliminary training phase is carried out using scene geometry. Specifically, all pixels above the horizon line are assumed to belong to non-road regions. Additionally, a rectangular region in front of the moving vehicle—corresponding to a quadrilateral in image coordinates—is identified as road-only, since it is unlikely to contain obstacles such as pedestrians or vehicles when the vehicle is moving at sufficient speed. The network is initially trained using these constraints while ignoring other regions. 

In the second stage, the model is refined through temporal consistency constraints. Corresponding feature points are extracted and matched for pairs of consecutive frames, ensuring that pixels representing the same physical location across frames receive consistent labels. This step mitigates temporal inconsistencies in segmentation and improves model robustness. 

The framework is model-agnostic and can be integrated with any segmentation architecture. In this study, we employ a lightweight CNN to enable efficient computation while maintaining high segmentation accuracy. The overall framework of the proposed approach is illustrated in figure \ref{figone}, which depicts both the initial training phase and the subsequent refinement using temporal consistency constraints. 
In summary, our contributions are as follows: 
\begin{itemize}
	\item We leverage scene geometry and camera calibration to adaptively generate weak labels, providing an initial structured signal for road vs. non-road segmentation. The weak labels for the road are adaptively adjusted based on the ego-vehicle's speed. 
	
	\item We introduce a mutual information–based temporal consistency loss that enforces stable predictions across consecutive frames by tracking local feature points.
	\item Our method achieves an IoU of 0.87 on the Cityscapes validation set, narrowing the gap to fine-tuned models (0.91 IoU), while requiring no labeled data.
	\item The framework is computationally lightweight, model-agnostic, and easily extendable to large-scale and dynamic autonomous driving scenarios.
\end{itemize}

\section{Related Work}
\label{Related Work}
\subsection{Unsupervised Semantic Segmentation}
\label{Unsupervised Semantic Segmentation}
Several recent methods~\citep{Yin2022top-down, wang2022discovery, Hamilton2022distilling, Seong2023hidden, Li2023Adaptive, Sick2024depth, Kim2024Expand} adopt a self-supervised pretrained ViT, such as DINO~\citep{caron2021emerging}, as their backbone. A prominent example is STEGO~\citep{Hamilton2022distilling}, which employs knowledge distillation to learn correspondences between features extracted from DINO and introduces a novel contrastive loss function.
There are also works that formulate unsupervised segmentation as a contrastive learning problem~\citep{Van2021proposals, Hamilton2022distilling, Seong2023hidden, Bao2025Context}. For instance, Van Gansbeke et al.~\citep{Van2021proposals} propose a method that pulls pixel embeddings belonging to the same object closer together while pushing embeddings from different objects further apart.
Beyond contrastive learning, another significant research direction~\citep{Lin2021superpixel, wang2022discovery, Melas2022localization, Aflalo2023Deepcut, Singh2023FODVid} explores spectral clustering for unsupervised semantic segmentation (USS). For example, FODVid~\citep{Singh2023FODVid} addresses unsupervised video object segmentation by using DINO features and optical flow to construct an adjacency graph, from which preliminary object masks are obtained via graph cuts.
While these approaches target general semantic segmentation, they are not specifically tailored to autonomous driving scenarios. In contrast, our method is explicitly designed for road scene videos. Moreover, it does not rely on a specific segmentation backbone and can be readily adapted to both lightweight models and more accurate, resource-demanding architectures.

\subsection{Unsupervised Video Segmentation}
\label{Unsupervised video Segmentation}
Despite image segmentation models being usable for videos by processing them frame-by-frame, doing this ignores valuable temporal information available in frame sequences. Research on unsupervised video segmentation has mainly focused on unsupervised video object segmentation (VOS).
AGS~\citep{Wang2023attention} 
links the near frames using a dynamic attention mechanism. The RTNet ~\citep{Ren2021Reciprocal} approach involves reciprocally transforming appearance features to motion features to mitigate the impact of misleading motion information on identifying primary objects. HFAN~\citep{Pei2022alignment} is a hierarchical feature alignment network that aligns appearance and motion features with primary object semantics. PMN~\citep{Lee2023prototype} generates component prototypes from RGB images and optical flow maps to capture both appearance and motion cues. 
MTNet~\citep{Zhuge2024temporal} incorporates a temporal transformer module to model long-range temporal dependencies and improve frame-to-frame interaction. These methods are designed for object-centric segmentation, relying on the motion of a target object across frames. In contrast, our method focuses on segmenting the road in driving-scene videos, where the target region is largely static and extends across the entire scene.


\subsection{Mutual Information For Unsupervised Semantic Segmentation}
\label{Mutual Information For Unsupervised Semantic Segmentation}
Mutual information maximization is a powerful principle in
discovering structure in visual data without manual annotations. 
One notable method is Invariant Information Clustering (IIC)~\citep{Ji2019IIC}. IIC maximizes mutual information between paired image representations—an image and its transformed version—to improve clustering performance while mitigating mode collapse. 
Another approach~\citep{Xiong2020vector} incorporates a mutual information loss within a joint framework that estimates scene depth, camera motion, and road segmentation. 
The work by Ouali et al.~\citep{Ouali2020Autoregressive} maximizes the mutual information between pairs of predictions obtained from two valid input orderings using different masked convolutions. InfoSeg~\citep{Harb2021Infoseg} segments images by maximizing the mutual information between local and global high-level features obtained from a self-supervised model. Again, these approaches target general segmentation tasks and do not exploit the geometric structure of road scenes. Nevertheless, to leverage temporal consistency, our approach makes use of the IIC loss.

\subsection{Road Segmentation with Minimum Supervision }
Recent research in road segmentation has increasingly focused on reducing or even eliminating the need for dense manual annotations. Weakly-supervised approaches rely on limited signals such as pseudo-labels or image-level tags~\citep{RobinetRefining2021,  KimWeaklydriving2024, Huangauto-driving2025}. Tsutsui et al.~\citep{Distantly} propose a distantly supervised road segmentation approach that exploits superpixel clustering guided by saliency maps. Several works have explored unsupervised segmentation in urban scenes, leveraging cues such as depth~\citep{Xiong2020vector, Sick2024depth} and cross-modal distillation~\citep{Drive-segment}. Another work by Tsutsui et al.~\citep{free-space} exploits a location prior and homogeneous texture, to generate pseudo-labels for free-space segmentation, assuming that pixels corresponding to free space are more likely to be located at the bottom and center of images captured by a front-facing camera.
Robinet et al.~\citep{co-teaching} build upon this approach and train a neural network to generalize past the label noise using Co-Teaching. Unlike Tsutsui et al.~\citep{free-space}, our proposed constraints are geometry-driven and deterministic, adapting explicitly to the ego-vehicle speed, rather than soft, appearance-based cues. Moreover, our method does not rely on clustering, making it inherently robust to errors arising from cluster assignments.
Overall, these approaches demonstrate that road segmentation can be learned with surprisingly little supervision, while leaving room for simpler and more robust geometry-driven solutions.

\section{Proposed Approach}
\label{Proposed Approach}

\begin{figure}[t]
	\centering
	\includegraphics[width=1\linewidth]{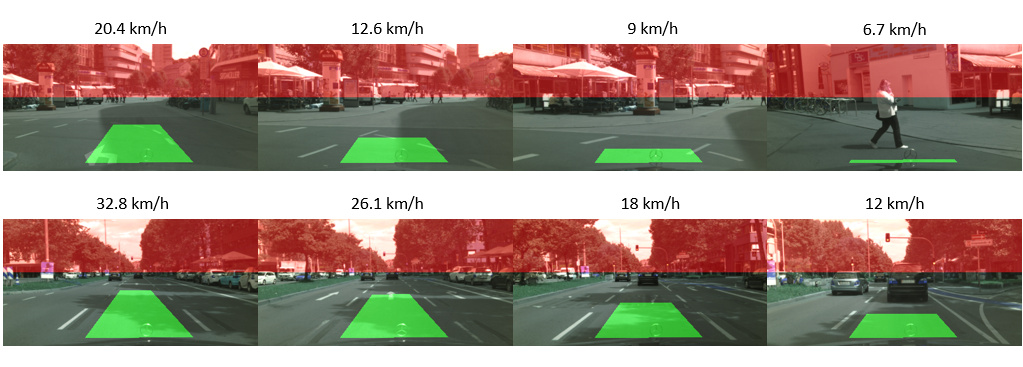}
	\caption{Qualitative examples of the predicted road region under different vehicle speeds. The green area denotes the estimated drivable road, while the red region corresponds to non-road areas above the horizon. }
	\label{figtwo}
\end{figure}

\subsection{Geometry-Driven Partial Masks}
\label{sec:partial-masks}
\textbf{Adaptive Road Region.} In the initial stage of training, we define a region in the image that almost certainly corresponds to the road. To do this, we consider a rectangular region on the ground whose width matches that of the car and extends from $X_A$ to $X_B$ meters in front of the vehicle (Fig.~\ref{figtwo}). Here, $X_A$ is the minimum distance at which the road becomes visible in the image, determined by the geometry of the ego vehicle and the camera placement. For the Cityscapes~\citep{cityscapes} dataset, this distance is 3.2 meters.  

The distance $X_B$ defines an adaptive safe distance that ensures no obstacles are present within the rectangle. It is computed based on the ego-vehicle's speed provided by the dataset. In many countries, a recommended safe following distance is proportional to vehicle speed; for example, one guideline suggests maintaining a distance of at least half the speed in meters (e.g., 40 meters at 80 km/h). We use this value to set $X_B$, ensuring that obstacles are unlikely to appear within the rectangular region. As a specific scenario, if a pedestrian crosses in front of the car, the vehicle speed  drops close to zero, causing the rectangle's length to shrink accordingly and reducing the chance of including pedestrians. The adaptive selection of this rectangular region is illustrated in Fig.~\ref{figtwo}.  

In Cityscapes~\citep{cityscapes}, the origin of the vehicle coordinate system lies on the ground directly beneath the center of the rear axle, with the X-axis pointing forward, the Y-axis pointing to the left, and the Z-axis pointing upward. The four 3D corner points of the rectangle are therefore $[L+X_A,~\pm W/2,~0]$ and $[L+X_B,~\pm W/2,~0]$ in vehicle coordinates, where $L$ is the distance from the rear axle to the front of the car, $W$ is the car width, and $X_A$ and $X_B$ are as defined above. These points are projected onto the image using the camera projection matrix, resulting in a quadrilateral region in image coordinates, shown as the green region in Fig.~\ref{figtwo}. This region reliably corresponds to the road; consequently, all pixels inside this quadrilateral are assumed to belong to the road.

\textbf{Non-Road Region.} Since the road never extends above the horizon line, we estimate the horizon by considering the line at infinity on the ground plane in vehicle coordinates. This line is projected into the image, and pixels above the projected line are assumed to be non-road. Given the intrinsic and extrinsic parameters, the horizon line can be projection into the image coordinates using the projection matrix $\mP = \mK [\mR, \t]$, where $\mK$ is the calibration matrix, and $\mR$ and $\t$ are the rotation matrix and translation vector of the camera with respect to vehicle coordinates. The projected horizon line in homogeneous coordinates is given by
\begin{equation}
	\mathbf{l} = (\mK \r_1) \times (\mK \r_2),
\end{equation}
where $\r_1$ and $\r_2$ are the first two columns of $\mR$, respectively.
\subsection{Initial Training with Geometric Constraints}
\label{sec:init_train}
The initial training phase enforces geometric constraints by optimizing a binary cross-entropy (BCE) loss defined only over the rectangular road region and the region above the horizon:
\begin{equation}
	\mathcal{L}_{\text{geometric}} =
	-\frac{1}{N_R} \sum_{i \in R} \log \sigma(y_i)
	-\frac{1}{N_H} \sum_{i \in H} \log \big(1 - \sigma(y_i)\big),
\end{equation}
where $y_i$ denotes the network output at pixel $i$, $\sigma(\cdot)$ is the sigmoid function, and $\sigma(y_i)$ represents the predicted probability that pixel $i$ belongs to the road.
The sets $R$ and $H$ correspond to the road and non-road partial masks defined in Sect.~\ref{sec:partial-masks} (shown as the green and red regions in Fig.~\ref{figtwo}), with $N_R$ and $N_H$ denoting the number of pixels in each region, respectively.  

By minimizing this loss, the network is encouraged to assign high probabilities to pixels consistent with the road mask $R$ and low probabilities to pixels consistent with the non-road mask $H$. This geometric pretraining provides a structured initialization, enabling the model to develop a reasonable understanding of the scene and facilitating more effective learning in subsequent training stages.

\subsection{Temporal Consistency and Feature Tracking}
\label{sec:temp_cons}
After the initial training phase, the model is refined by enforcing temporal consistency across video frames. First, feature points are extracted using the Shi-Tomasi corner detection algorithm~\citep{Shi1994Track}. Then these points are tracked across consecutive frames using the Lucas-Kanade optical flow method~\citep{LK1981Tracker}. We then refine the training process by forcing the network to assign identical labels to a feature point and its tracked correspondence across frames, using mutual information maximization via the IIC loss \citep{Ji2019IIC}. Consider a set of feature point locations $\x_1, \x_2, \ldots, \x_n$ in image 
$I$ and their tracked counterparts $\x_1', \x_2', \ldots, \x_n'$ in image $I'$. 
Let $y_i$ and $y_i'$ denote the network output logits on $I$ and $I'$ at pixel locations $\x_i$ and $\x_i'$, respectively. Let $\s_i = [1{-}\sigma(y_i), \sigma(y_i)]^T$ and $\s_i' = [1{-}\sigma(y_i'), \sigma(y_i')]^T$ be the vectors of class probabilities, where $\sigma(\cdot)$ represents the sigmoid function. To compute the IIC loss, we first obtain the joint distribution matrix 
\begin{equation}
	\mP = \frac{1}{n} \sum_{i=1}^{n} \s_i\,(\s_i')^T \in \mathbb{R}^{2 \times 2}.
\end{equation}

Then we compute the marginal distributions $\p, \p' \in \mathbb{R}^2$ by summing over the rows and columns of $\mP$, respectively. 
Now, the IIC loss \citep{Ji2019IIC} can be evaluated by substituting the joint probability matrix 
$\mP$ into the mutual information expression:
\begin{equation}
	\label{eq:iic_loss}
	\mathcal{L}_{\text{Consistency}} = -I(\mP) = -\sum_{i=1}^2 \sum_{j=1}^2 \mP_{ij} \cdot \ln \frac{\mP_{ij}}{p_i \cdot p_{j}'} .
\end{equation}




The final loss function combines the initial geometric constraints with the temporal consistency loss:
\begin{equation}
	\mathcal{L}_{\text{Final}} = \mathcal{L}_{\text{Geometric}} +  \mathcal{L}_{\text{Consistency}}.
\end{equation}
This ensures that the network not only respects the initial geometric priors but also produces temporally coherent predictions across video sequences.

\subsection{Hard Negative Mining}
\label{sec:HNM}
\begin{figure}[t]
	\centering
	\includegraphics[width=0.6\linewidth]{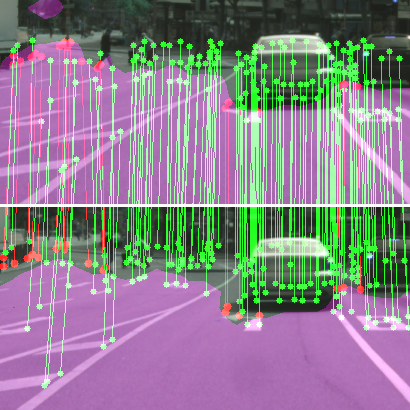}
	\caption{A pair of frames from the Cityscapes dataset with its segmentation mask and point tracking results. The purple area represents the road predicted at the 10th training epoch. Pairs of tracked points with consistent (green) and inconsistent (red) labels predicted by the network are also drawn. Only the inconsistent pairs are kept for training.}
	\label{fig:hard_negative}
\end{figure}

In the refinement stage, inspired by \emph{hard negative mining}, we focus training on temporally inconsistent feature correspondences to sharpen the model’s decision boundaries. Rather than applying the consistency loss~\eqref{eq:iic_loss} to all pairs of tracked features, we maintain a pool of candidate pairs. After each epoch, pairs of tracked points whose predicted labels differ across consecutive frames are identified as inconsistent (see Fig.~\ref{fig:hard_negative}) and added to this pool.  

Each epoch consists of two steps: (1) extracting new inconsistent correspondences based on the current model predictions, and (2) retraining the network using the updated pool of candidate pairs to improve segmentation accuracy. This iterative procedure continues until temporal inconsistencies are reduced, resulting in a stable and robust segmentation model.

\section{Experiments}
\label{Experiments}
\subsection{Dataset}
To evaluate the performance of our method, we utilized the Cityscapes dataset ~\citep{cityscapes}, which offers a comprehensive collection of urban driving scenes captured in German cities and is widely used for benchmarking semantic segmentation in autonomous driving. 
For training, we employed 2,900 sequential frames from the Cityscapes dataset. 
Depending on the evaluation protocol of the baseline methods, we use both the Cityscapes test set and the validation set, comprising 1,525 and 500 images, respectively.

\subsection{Implementation Details}
Our method is loss-driven and therefore agnostic to the choice of segmentation network architecture. In principle, it can be applied to any segmentation model. However, to ensure a fair comparison, we deliberately selected a network that is lighter than those used in all baseline methods.  

Specifically, we adopt a \emph{Lite R-ASPP} architecture with a MobileNetV3-Large backbone~\citep{Howard2019net}, which is known for its lightweight design, computational efficiency, and strong performance in semantic segmentation tasks. The model is trained from scratch using a two-phase training scheme on a single NVIDIA T4 GPU. The best-performing weights are selected based on validation set performance.  
In the first training phase (Sect.~\ref{sec:init_train}), the network is trained for 100 epochs on 2900 frames from the Cityscapes dataset. In the second phase, the loss function is augmented with the temporal consistency term (Sect.~\ref{sec:temp_cons}), applied to 870 pairs of frames each temporally separated by either 5 or 10 frames. This phase is trained for 100 epochs and requires approximately 300 minutes. We evaluate this stage both with and without hard negative mining (Sect.~\ref{sec:HNM}).  

In both training phases, the initial learning rate is set to $10^{-4}$, and the Adam optimizer is used to ensure stable convergence.


\subsection{Evaluation}
The performance of the proposed method was assessed using the IoU metric, which is widely regarded as the most reliable measure of segmentation accuracy.
For fairness, we compare our approach only against methods that perform binary segmentation and whose primary objective is road segmentation, excluding approaches designed for multi-class segmentation. Notice that all the baseline methods use larger segmentation models than ours. 

Table \ref{tableone} reports a quantitative comparison between our proposed method and several existing approaches on the Cityscapes validation set. The compared methods achieve IoU scores in the range of 0.80 to 0.83, reflecting the performance of prior work under similar evaluation settings. Our method attains an IoU of 0.87, providing a clear improvement over all listed baselines. 
We also compare our unsupervised approach with a baseline based on the same LR-ASPP architecture, initialized with weights pretrained on a subset of the COCO dataset~\citep{Coco} and subsequently adapted to Cityscapes using our unsupervised framework. This baseline achieved an IoU of 0.91, while our fully unsupervised method reached 0.87 on the Cityscapes validation set. The modest gap of 0.04 highlights the effectiveness of our fully unsupervised framework, especially considering that it requires no manual labeling.

\begin{table}[t]
	\centering
	\begin{tabular}{l c r}
			\toprule
			Method & Annotation & IoU \\
			\midrule
			Distant Supervision~\citep{Distantly} & image level& 0.80 \\	
			Minimized Supervision ~\citep{free-space} &none& 0.83 \\
			Joint Learning~\citep{Xiong2020vector} &none& 0.83  \\
			Stochastic Co-Teaching~\citep{co-teaching} &none& 0.82 \\
			\midrule
			Unsupervised Fine-Tuning&pixel level& 0.91 \\		
			Unsupervised Method (Ours) &none& \textbf{0.87} \\
			\bottomrule
		\end{tabular}
		\caption{Performance comparison of the proposed unsupervised method with existing approaches on the Cityscapes validation set.}
		\label{tableone}
	\end{table}

		To further evaluate the effectiveness of the proposed method, we compare its performance with the weakly-supervised free-space segmentation method~\citep{free-space}, which achieved strong results on road segmentation for the test set. As shown in Table \ref{tabletwo}, our unsupervised framework achieves slightly higher IoU than the weakly-supervised baseline. This improvement stems from the nature of our constraints: our approach employs geometry-driven, deterministic constraints that explicitly adapt to the ego-vehicle speed. In addition, our method does not depend on clustering, making it inherently more robust to errors caused by incorrect clustering. Despite its conceptual simplicity, this design yields a strong and efficient baseline for unsupervised road segmentation.

		\begin{table}[t]
			\centering
			\begin{tabular}{l c r}
				\toprule
				Method & Supervision & IoU \\
				\midrule
				Minimized Supervision ~\citep{free-space} &none& 0.85 \\	
				Unsupervised Method (Ours) &none& 0.86 \\
				\bottomrule
			\end{tabular}
			\caption{Performance comparison of the proposed unsupervised method with existing approaches on the Cityscapes test set.}
			\label{tabletwo}
		\end{table}

\begin{figure*}
			\centering
			\includegraphics[width=1\linewidth]{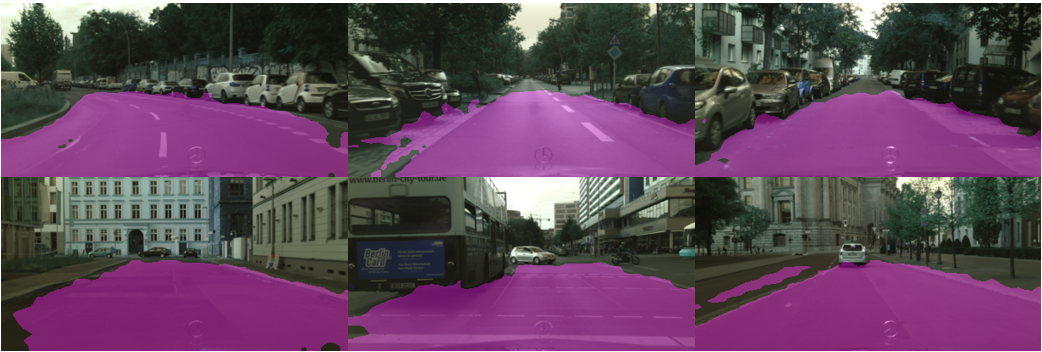}
			\caption{Examples of segmented Cityscapes images using the presented method}
			\label{figfour}
		\end{figure*}

		\noindent\textbf{Visual Analysis and Interpretation.} 
		The qualitative results of the segmentation outputs further illustrate the effectiveness of the proposed method. Fig.~\ref{figfour} presents several examples of road segmentation outputs on images from Cityscapes. The model successfully delineates drivable areas with notable precision, effectively distinguishing between road surfaces and non-drivable regions. The illustrated results highlight the model's ability to adapt to varying road textures, lighting conditions, and perspectives. 
		The robustness of the proposed method is attributed to its two-stage training approach. First, geometric constraints provide an initial structured learning signal, ensuring that the network understands fundamental road structures. Second, temporal consistency loss refines segmentation by leveraging frame-to-frame feature correspondence, leading to smoother and more coherent predictions over time.
		Minor inaccuracies are observed in challenging areas, such as vehicle tires and sidewalks, where the lack of explicit supervision may lead to occasional misclassifications, which could be further improved by integrating a small amount of weak annotations in future research. Nonetheless, the model consistently captures the general road structure with high fidelity.
		The results suggest that unsupervised learning with geometric and temporal cues is a promising direction for road segmentation in autonomous driving. 
		
		\subsection{Ablation Study}
		Table \ref{tablethree} analyzes the contribution of temporal consistency and hard negative mining to the proposed framework. In the first training phase, the network was trained solely using geometric constraints as supervisory signals via partial masks, achieving an IoU of 0.74. This baseline indicates that geometric priors alone provide a reasonable initialization but are insufficient to fully capture appearance variability and dynamic scene content. In the second phase, temporal consistency is introduced in addition to the geometric constraints, while hard negative mining is disabled and all consistent and inconsistent feature points are used for training. This setting improves the IoU to 0.85, demonstrating that enforcing temporal coherence across consecutive frames complements geometric supervision. Finally, incorporating hard negative mining  further increases performance to an IoU of 0.87, while reducing the training time by half. By focusing the learning process on challenging inconsistent samples and filtering out easy or redundant points, hard negative mining accelerates convergence and improves robustness.
		
		\begin{table}[t]
			\centering
			\begin{tabular}{l c r}
				\toprule
				Method & IoU \\
				\midrule
				Geometric Constraints (no Temporal Consistency) & 0.74 \\
				Temporal Consistency (no Hard Negative Mining) & 0.85 \\
				Temporal Consistency + Hard Negative Mining & 0.87 \\
				\bottomrule
			\end{tabular}
			\caption{Influence of temporal consistency and of hard negative mining on the Cityscapes validation set.}
			\label{tablethree}
		\end{table}

\section{Conclusion}
\label{Conclusion}
This study introduced a novel, fully unsupervised framework for road segmentation using a single monocular camera. The method leverages scene geometry and temporal consistency constraints to train a segmentation network without requiring any labeled data. Our method operates purely on the loss function and is therefore agnostic to the underlying segmentation architecture. Trained sequentially with geometric and temporal consistency objectives, the model produces stable and accurate segmentation on unseen Cityscapes test data. This performance demonstrates that strong results can be obtained with a lightweight CNN while entirely eliminating manual annotation. Compared to existing unsupervised approaches, the proposed framework is computationally efficient, relying on simple geometric priors and point tracking rather than costly pseudo-labeling, and exhibits strong generalization beyond the dataset-specific conditions. Future work could further enhance performance by integrating weak supervision or domain adaptation strategies to improve robustness across diverse road environments

\section*{CRediT authorship contribution statement}

\textbf{Sara Hatami Rostami:} Conceptualization, Methodology, Software, Formal analysis, Investigation, Visualization, Writing – original draft.

\textbf{Behrooz Nasihatkon:} Supervision, Conceptualization, Methodology, Investigation, Writing – review \& editing.

\section*{Declaration of competing interest}

The authors declare that they have no known competing financial interests or personal relationships that could have appeared to influence the work reported in this paper.




\end{document}